\documentclass[10pt,twocolumn,letterpaper]{article}

\usepackage{iccv}
\usepackage{times}
\usepackage{epsfig}
\usepackage{graphicx}
\usepackage{amsmath}
\usepackage{amssymb}
\usepackage{todonotes}

\DeclareMathOperator*{\argmax}{\arg\!\max}

\usepackage[pagebackref=true,breaklinks=true,letterpaper=true,colorlinks,bookmarks=false]{hyperref}

 \iccvfinalcopy 

\ificcvfinal\fi
\begin{document}

\title{Semantic Pose using Deep Networks Trained on Synthetic RGB-D}

\author{Jeremie Papon and Markus Schoeler\\
Bernstein Center for Computational Neuroscience (BCCN)\\
III. Physikalisches Institut - Biophysik, Georg-August University of G\"{o}ttingen \\
{\tt\small jpapon@gmail.com mschoeler@gwdg.de}
}

\maketitle

\begin{abstract}
  In this work we address the problem of indoor scene understanding from RGB-D images. Specifically, we propose to find instances of common furniture classes, their spatial extent, and their pose with respect to generalized class models. To accomplish this, we use a deep, wide, multi-output convolutional neural network (CNN) that predicts class, pose, and location of possible objects simultaneously. To overcome the lack of large annotated RGB-D training sets (especially those with pose), we use an on-the-fly rendering pipeline that generates realistic cluttered room scenes in parallel to training. We then perform transfer learning on the relatively small amount of publicly available annotated RGB-D data, and find that our model is able to successfully annotate even highly challenging real scenes. Importantly, our trained network is able to understand noisy and sparse observations of highly cluttered scenes with a remarkable degree of accuracy, inferring class and pose from a very limited set of cues. Additionally, our neural network is only moderately deep and computes class, pose and position in tandem, so the overall run-time is significantly faster than existing methods, estimating all output parameters simultaneously in parallel on a GPU in seconds. 
\end{abstract}

\section{Introduction}
In order for autonomous systems to move out of the controlled confines of labs, they must acquire the ability to understand the cluttered indoor environments they will inevitably encounter. While many researchers have addressed the problems of pose estimation, object detection, semantic segmentation, and object classification separately, comprehensive understanding of scenes remains an elusive goal. To this end, in this work we propose an architecture which is able to perform all of the above tasks in concert using a single artificial neural network. 

Classification in cluttered indoor scenes can be extremely challenging, especially when trying to classify instances of objects which have never been observed before. Considering only 2D color information only compounds this problem, as clutter can easily cause vast changes in the visible signature of otherwise distinguishable items. 3D geometric features, on the other hand, tend to be less susceptible to clutter and have (especially for furniture) geometric features which generalize well across the class. As such, in this work we use 3D geometric features in addition to standard RGB channels.

\begin{figure}[t]
  \centering
  \includegraphics[width=0.5\textwidth]{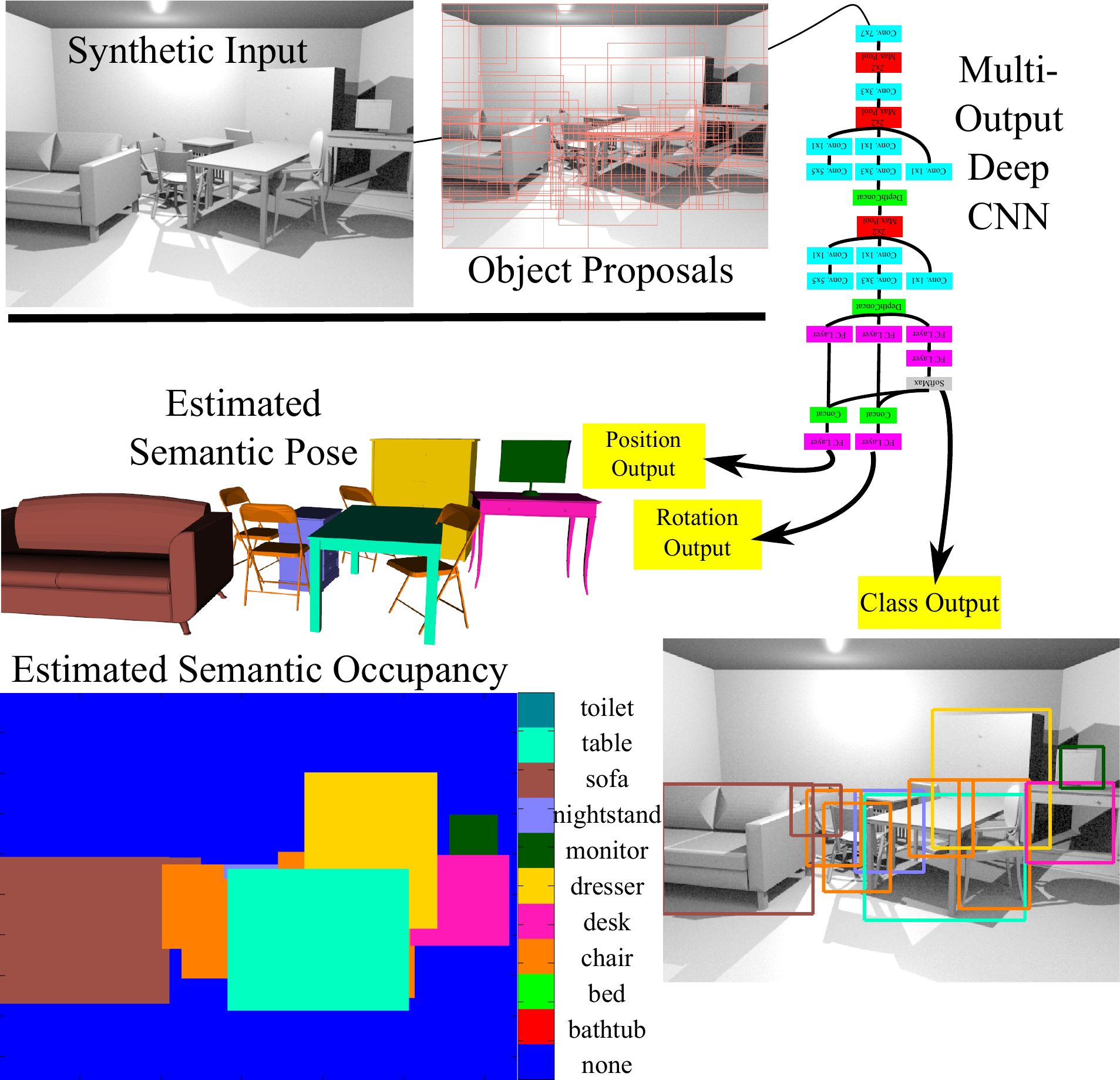}
  \caption{Overview of our approach. Normals for a scene are efficiently calculated using \cite{Holzer12}, proposals are generated using \cite{Krahenbul14}, and then fed through our synthetically trained CNN. Outputs are then consolidated using non-maximum suppression, leading to a scene class \& pose heat map and a scene rendered with generalized models.}
  \label{fig:overview}
\end{figure}

\begin{figure}[t]
  \centering
  \includegraphics[width=0.5\textwidth]{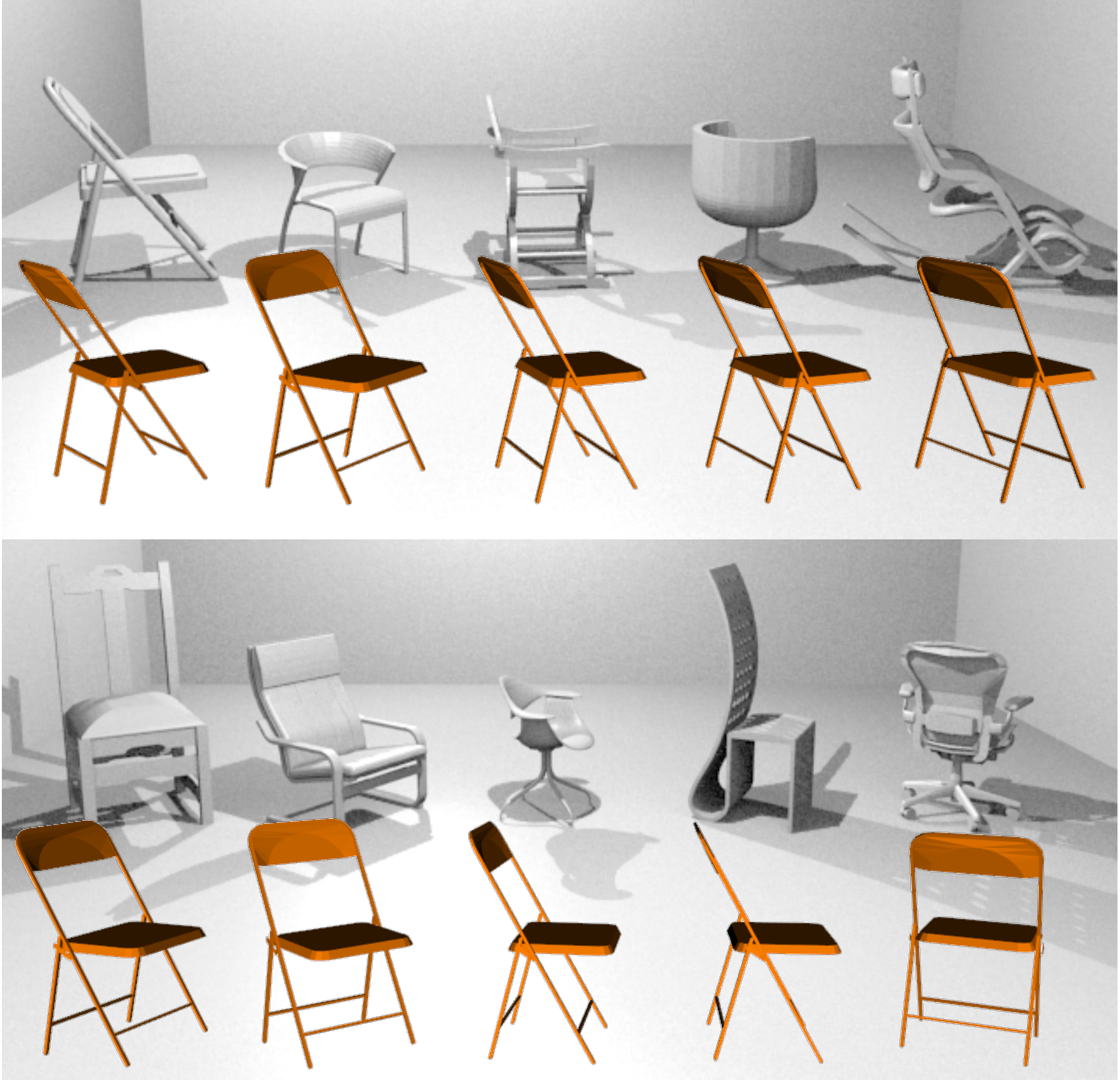}
  \caption{Example of estimated pose output (overlaid as a generic orange model) for chairs from the test set. Pose here is shown using a generic chair model. None of these test models were observed in training.}
  \label{fig:qualitative_chairs}
\end{figure}

Pose estimation in-the-wild is another difficult problem, as it requires estimating pose for object instances which have never been observed before. For example, consider the task of helping a human to sit down in a chair - to be of any help, one must be able to determine pose of the back-rest, the seat area, and the supporting legs - even on types of chairs that one has never seen before. In this work we will show that just such a task is possible, to a surprising degree of accuracy, using a wide, deep, multi-stage CNN trained on synthetic models. In fact, it is possible to do so even with wholly unobserved types of chairs - for example, in Fig.~\ref{fig:qualitative_chairs}, none of the chair models were seen in training. Moreover, we shall demonstrate that it is possible to estimate such poses even in complex cluttered scenes containing many classes of furniture (\eg see Fig. \ref{fig:synthetic_examples}).

Our approach, outlined in Fig. \ref{fig:overview}, uses a relatively complex CNN architecture to solve our three sub-tasks; class-, pose-, and position-estimation of objects, concurrently. One unusual aspect of our network is that it recombines class output back into the network layers which calculate pose and position, allowing the network to accurately determine pose for multiple classes within a single architecture. Furthermore, we are able to train this large network by using synthetic rendered RGB-D scenes consisting of randomly placed instances from a dataset of thousands of 3D object models. Our training scenes are generated on the fly on the CPU and a secondary GPU as we train on the primary GPU, allowing us to have a training set of virtually unlimited size at a completely hidden computational cost. Finally, we use a small number of transfer learning iterations using a small set of real annotated images to adapt our network to the modality of real indoor RGB-D scenes. 

To demonstrate the effectiveness of our approach, we perform a variety of experiments on both synthetic and real scenes. Our pose estimation and classification results outperform existing methods on a difficult real dataset. We also present qualitative and quantitative results on both real and synthetic data which demonstrate the capability of our system to distill semantic understanding of scenes. Moreover, we do these tasks jointly in a single forward pass through our network, allowing us to produce results significantly faster than existing methods.


\subsection{Related Work}
As we propose to solve multiple problems in tandem in this work, there is a substantial body of work which could be considered related. We will restrict ourselves to those recent works which deal exclusively with RGB-D data and/or use CNNs to accomplish one or more of our sub-tasks.

As a first step in a pipeline to parse full scenes, the image is typically broken down into small ``object proposals'' to be considered by other methods. For example, in Silberman \etal \cite{Silberman12} they perform an over-segmentation, and then iteratively merge regions using classifiers which predict whether regions belong to the same object instance. These are then classified using an ensemble of features with a logistic regression classifier. 

Couprie \etal \cite{Couprie13} take a different approach, instead using a multi-scale CNN to classify the full image, and then use superpixels to aggregate and smooth prediction outputs. While this allows them to extract a per-pixel semantic segmentation, they fail to achieve very high scores in important classes, such as table and chair. 
Hariharan \etal \cite{Hariharan14} also predict pixel-level class associations, but classify region proposals instead of the full image. They also use a CNN as a feature extractor on these regions, before classifying into categories with an SVM and aggregating onto a coarse mask. They then use a second classifier stage on this coarse mask projected on to superpixels to extract a detailed segmentation.  
While these results are interesting, we question the overall utility of such a fine grained segmentation, as it does not provide pose with respect to a class-level representation. 

Song and Xiao \cite{Song14} use renderings of 3D models from many viewpoints to obtain synthetic depth maps for training an ensemble of Exemplar-SVM classifiers. They use a 3D sliding window to obtain proposals during testing and perform non-maximum suppression to obtain bounding boxes. While this 3D sliding window approach is able to handle occlusions and cluttered scenes well, it is very expensive (tens of minutes per image), requiring testing of many windows on many separate detector classifiers. 

Guo and Hoiem \cite{Guo13} predict support surfaces (such as tables and desks) in single view RGB-D images using a bottom up approach which aggregates low-level features (\eg edges, voxel occupancy). These features are used to propose planar surfaces, which are then classified using a linear SVM. While they provide object-class pose annotations for the NYUv2 set which we use in this paper, they do not classify objects or their pose themselves.

Object detection in RGB-D is addressed directly by Gupta \etal \cite{Gupta14} using a CNN which classifies bounding-box proposals in a room-centric embedding. As with other approaches, they use superpixels to aggregate their classifier results in order to get class instance segmentations. Lin \etal \cite{Lin13} use candidate cuboids, rather than bounding boxes, and classify them using a CRF approach. While they achieve good overall classification performance, they merge similar classes (such as table and desk), and while their cuboids give them spatial extent of objects, they do not give pose.

In contrast to the above methods, we do not need expensive and difficult to obtain annotated ground truth data for training. Instead, we use synthetic renderings of scenes containing 3D models pulled from the Internet. While these models need to be aligned to a common pose, this is a relatively inexpensive operation which has already been performed in the ModelNet10 database \cite{Wu15-3DShapenets}. 

The only other work to address pose directly, that of Gupta \etal~\cite{Gupta15}, suffers from using unrealistic training data - training instances are single objects rendered in empty space. In contrast, our synthetic data is cluttered and contains realistic noise, as we use a camera model which closely replicates Kinect-like sensors. Because of this, our trained networks are far more effective on real data - we test on the full NYU dataset, while they must leave out instances that have many ($>$50\%) missing depth pixels. Additionally, since we work with full scenes rather than single object instances, our model is trained on and can thus handle inter-object occlusions, rather than only self-occlusions. Moreover, their network contains separate top-level layers for each object class, while we only need a single output network for pose for all classes. Their method is also computationally demanding, requiring about a minute per image per class, while ours runs in a few seconds for all classes.

\section{Synthetic RGB-D Scenes}
One of the main obstacles to using deep CNNs on RGB-D data is the lack of large annotated datasets. This is especially true for pose data, where annotation of a set of the size required for training a deep network is simply not feasible. Synthetic data, on the other hand, provides labeled segmentations and exact pose for free, but has yet to find widespread use, likely owning to the difficulty of rendering photo-realistic scenes. Fortunately, RGB-D data lends itself to the use of synthetic data due to the simplicity with which depth data can be rendered realistically. One only needs to simulate the active model of the sensor, and can largely ignore lighting, textures, and surface composition. 

\begin{figure}[t]
  \centering
  \includegraphics[width=0.45\textwidth]{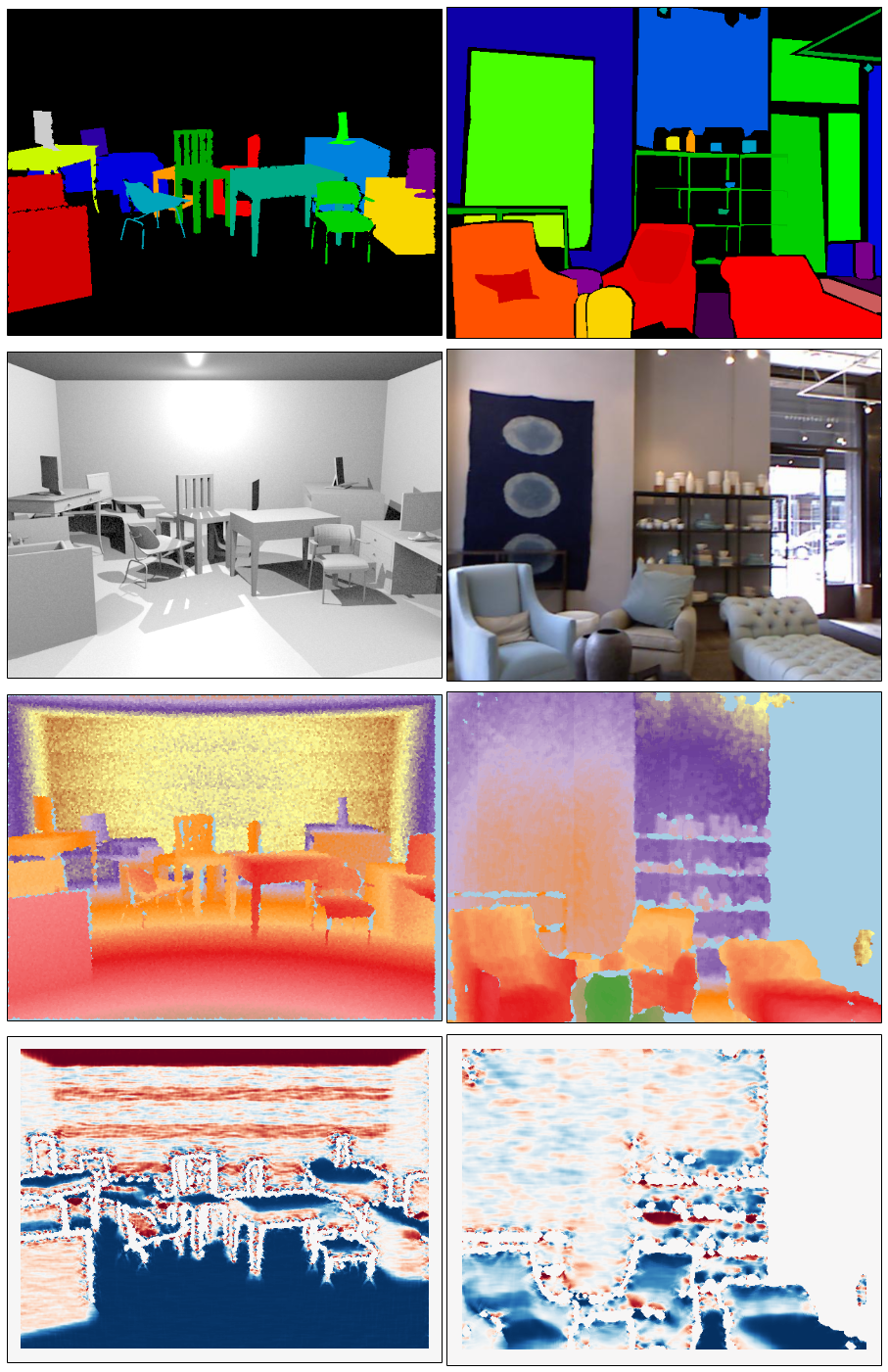}
  \caption{Example of a randomly generated synthetic scene using our rendering pipeline (left) and a scene from the NYUv2 dataset. The rows show A. Ground truth labels, B. RGB Channel, C. Depth Channel, D. Normals calculated using \cite{Holzer12}. The left column shows our synthetic data, and the right an image from NYUv2 \cite{Silberman12}.}
  \label{fig:synthetic_examples}
\end{figure}

Our synthetic scenes are produced by sequentially placing objects models at random in a virtual room. As each object is placed, we ensure that its mesh does not intersect with other objects or the room surfaces. Additionally, we use context cues to increase the realism of our scenes - large furniture (\eg sofas or beds) is biased to occur near walls, chairs are biased to occur near tables and desks, and monitors are always placed on top of desks. We also randomly place a light source on the ceiling in the room to simulate shadow effects in the rendered intensity images. An example random scene is shown in Fig.~\ref{fig:synthetic_examples}. We have published the dataset used in this work for use by the community, and have also included the code for easily generating more scenes on the fly at training time\footnote{--Website removed for blind review--}. 

\subsection{Rendering \& Camera Model}
We build upon the BlenSor sensor simulation toolbox \cite{Gschwandtner11} to generate realistic RGB-D renderings of our randomly generated scenes. The ray-tracing used allows us to reproduce the real geometry of the Kinect sensor, faithfully simulating the projection of an IR pattern onto the scene and observation of the returns. As Kinect-type sensors will generally fail when reflections are present, we can safely limit our ray-tracing to a single hop. Additionally, we simulate the 9x9 correlation window required by the Kinect to produce depth measurements \cite{Smisek13} and add Perlin noise to the disparity measurements. We also use a standard Blender pipeline to render accompanying RGB images, though these are not photo-realistic due to a lack of textures on the object models and a simplified lighting model. As we only use the intensity channel, we found this simple RGB rendering to be sufficient, especially given that we use transfer learning to adapt to real sensor images.

\subsection{Models}
Our models must be aligned to a reference pose, preventing us from simply pulling CAD models from the Internet. Fortunately, the Princeton ModelNet10 dataset \cite{Wu15-3DShapenets} provides a varied set of pose-aligned models for ten object categories: bathtub, bed, chair, desk, dresser, monitor, nightstand, sofa, table, and toilet. We use the standard training/testing split provided. As the models are not scale-normalized, we choose a reasonable range of values per class, and rescale models randomly to fall within these ranges. Models are inserted on the floor or a supporting surface of our synthetic rooms at random locations with random rotations around the axis perpendicular to the floor. 

\section{Network Architecture}
We tested several different network configurations, all of which involved at least two Krizhevsky-style \cite{Krizhevsky12} (\ie Conv-ReLU-Pooling) convolutional layers at the input. Our most successful model, shown in Fig.~\ref{fig:network_architecture}, then uses a succession of Network-in-Network (NiN) layers \cite{Lin14}, in a configuration similar to the recent ``Inception'' architecture \cite{Szegedy14}. We then use separate multilayer perceptrons with two hidden layers to classify. Additionally, we connect our class output back into the second hidden layer of our pose and position classifiers.

\subsection{Input Preprocessing}
The input to our network is a 96x96 real-valued image consisting of five layers - an intensity layer, a depth layer, and three layers representing the surface normal vector (\eg $(normal_x, normal_y, normal_z)$). Depth values are used directly (in meters) and intensity values are computed from RGB using CIE 1931 linear luminance coefficients. While hue information is likely useful, our synthetic models are not colored, so we chose not to use it. We exploit the structured nature of RGB-D data to efficiently compute surface normals using the method of Holzer \etal \cite{Holzer12}. All channels are zero centered using mean values computed on a random sample of proposed bounding boxes from our training set. 

\subsection{Proposal Generation}
\begin{figure}[t]
  \centering
  \includegraphics[width=0.45\textwidth]{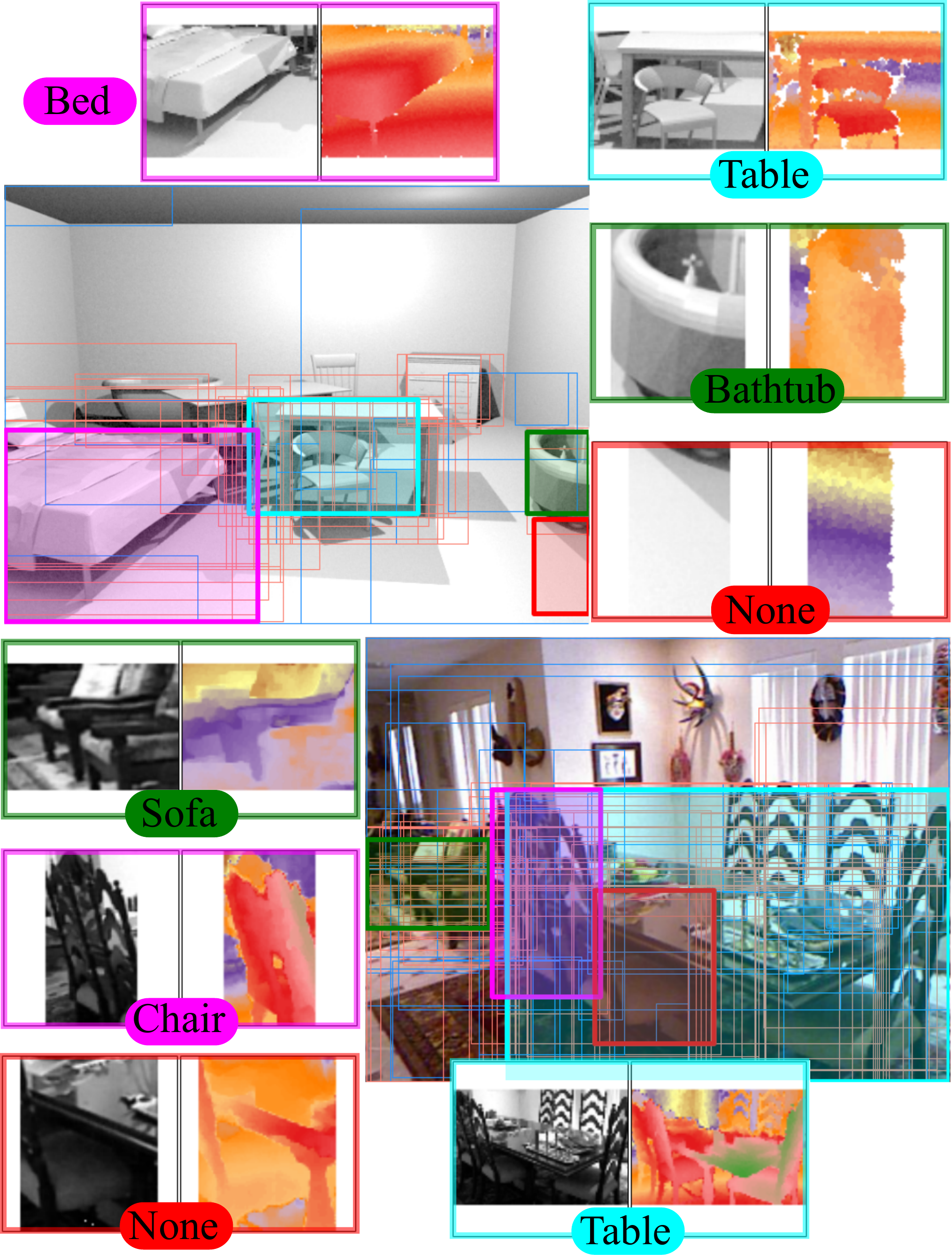}
  \caption{Example of bounding box proposals on synthetic data (top) and the NYUv2 Dataset\cite{Silberman12} (bottom).}
  \label{fig:bounding_boxes}
\end{figure}

Bounding box proposals are generated using the Geodesic Object Proposals (GOP) of Krähenbühl and Koltun \cite{Krahenbul14}. The method identifies level sets in geodesic distance transforms for seed points which are placed using classifiers optimized for object discovery. The method produces accurate and consistent bounding boxes at a low computational cost (approx. 1 second per image). Examples of proposed bounding boxes on our synthetic rendered images as well as on the NYUv2 images are shown in Fig.~\ref{fig:bounding_boxes}. We do not consider depth when generating our proposals, as we did not find it to be helpful in practice - a result supported by other researchers \cite{Silberman12}.  

\subsection{Network Layers}
\begin{figure}[t]
  \centering
  \includegraphics[width=0.25\textwidth]{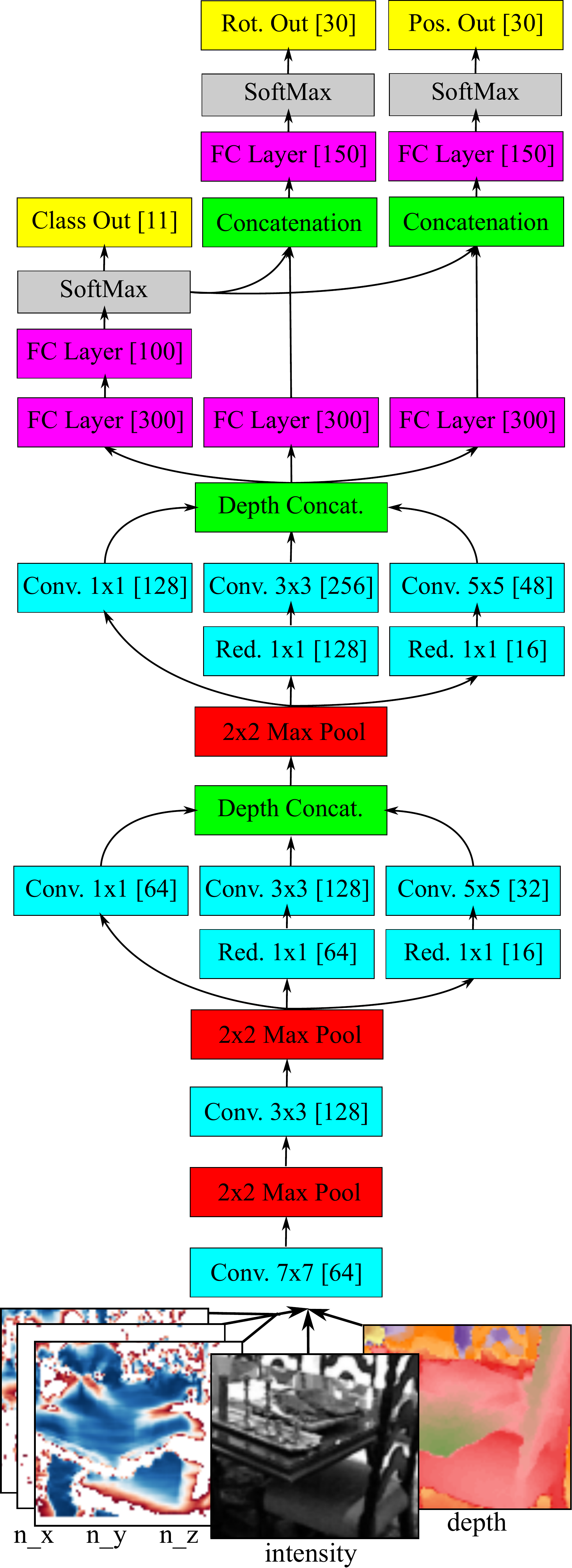}
  \caption{Network architecture of our most successful model. Numbers in brackets are either number of filters (conv. layers) or nodes (FC layers). The input consists of 96x96 5 channel images with normals, intensity, and depth. }
  \label{fig:network_architecture}
\end{figure}

We tested four models in total: two ``standard'' Krizhevsky-style CNNs, and two larger neworks with ``inception''-style layers. The first, baseline, model is a standard CNN network closely resembling the successful model of Krizhevsky \etal \cite{Krizhevsky12} - it consisted of five Conv-ReLU-Pooling layers, followed by two fully-connected (FC) classification layers for each output layer. The second model takes the class output and reconnects it back into the fully connected layers for pose and depth estimation. The third model expands the network by replacing the top convolutional layers with two inception-style\cite{Szegedy14} network-in-network layers. Lastly, the largest model increases the number of nodes even further by adding another inception layer, as well as an additional FC multi-layer network branching off from the first inception layer and reconnecting as an additional input to the classification FC layers. Dropout was used on the convolutional layers as well as the fully connected (FC) layers of the perceptrons in all models to limit over-fitting. Our most successful model is shown in Fig.\ref{fig:network_architecture}. 

\section{Training}
We train our networks to predict three outputs: a class label, a rotation around the floor normal axis, and a distance from the camera. Combined with a bounding box in the image plane, these allow us to generate a full description of the pose of furniture with respect to the set of standard reference models used by Guo and Hoeim \cite{Guo13}. 
We chose to predict binned rotation and depth values rather than perform a regression as we found that, in practice, the training was much more stable for classification, even with the relatively large number of bins ($n=30$, \ie 12 degrees per bin) used. 
We use a standard SoftMax cross-entropy loss for the class output, but adopt a soft-binning scheme for the pose and depth outputs. This takes a weighted (by $\gamma$) average of the local bins around the ground truth in the loss function, in order to help with poses which lie near bin boundaries: 
\begin{equation}
  L_i = -\log\left(\frac{\sum_{k=-1}^{1} \gamma_{i+k} e^{f_{y_i}}}{ \sum_j e^{f_j} }\right) : \sum_{k=-1}^{1} \gamma_{i+k} = 1 .
\end{equation}

\subsection{Synthetic Data}
While we can generate unlimited data at training time, for comparison purposes we trained on a fixed set of 7000 randomly generated scenes, consisting of a total of 59784 instances from our set of 2842 pose-aligned models from the ModelNet10 dataset \cite{Wu15-3DShapenets}. There is no constraint on the number of synthetic scenes possible - we only limited ourselves due to time constraints and in order to compare models. Our validation set was generated randomly during training. Additionally, we generated a test set of 1000 random scenes, using a separate set of 812 models from the same dataset. For training, we extracted bounding boxes using GOP and selected those that had 70\% overlap with the ground truth, leading to a total of 300,000 training instances. We scale bounding box proposals to fit our input size by fitting the larger dimension to our window size and zero padding the other.
\begin{figure*}[t]
  \centering
  \includegraphics[width=1.0\textwidth]{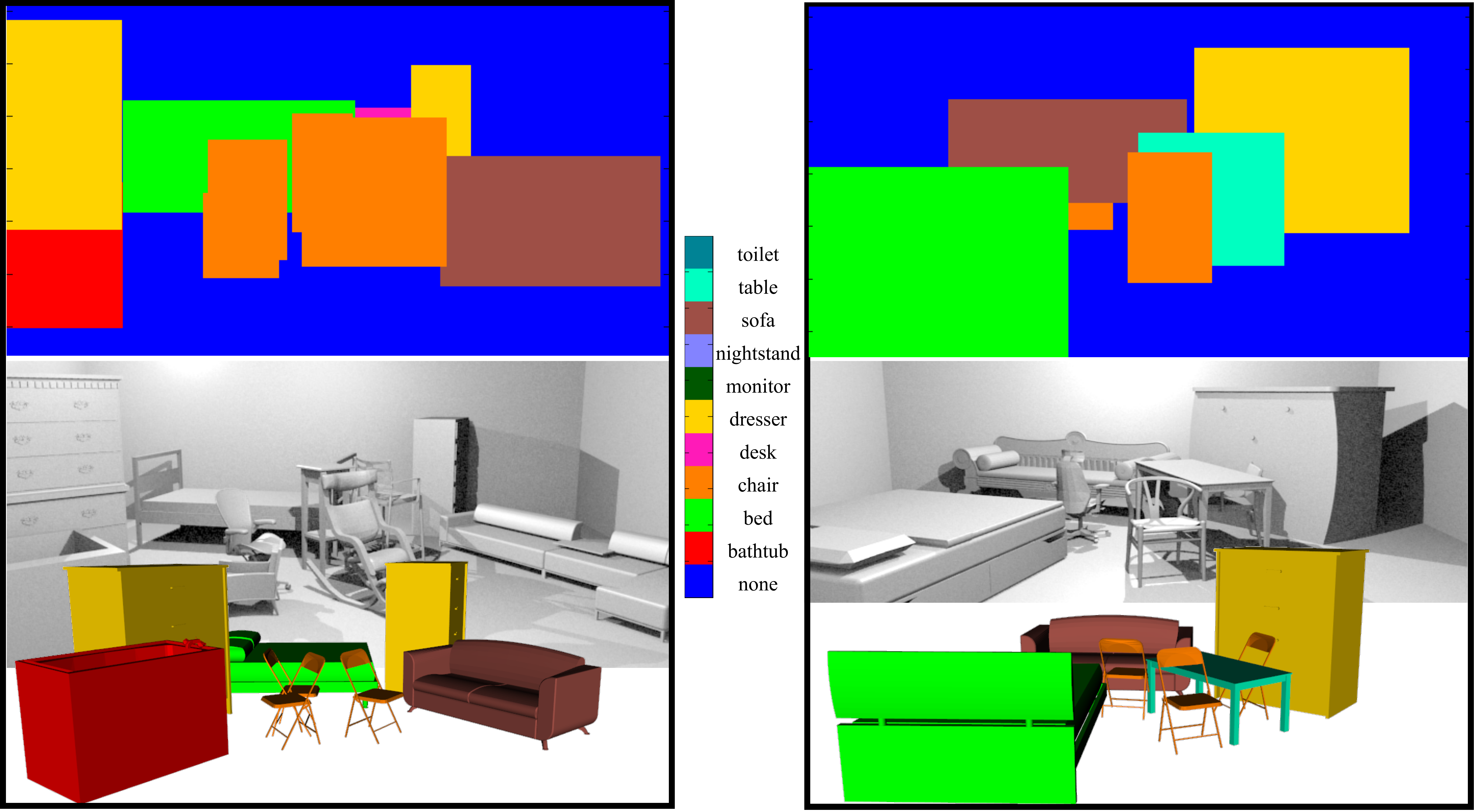}
  \caption{Qualitative pose and classification results from our synthetic test set. The top row shows our estimated semantic heatmap, while the bottom row shows pose and classification using generic models. The models in the test set are distinct from those used in training - this means that poses here are general class-based pose, rather than specific model-based.}
  \label{fig:qualitative_synthetic}
\end{figure*}

Additionally, we randomly selected an equal number of ``none''-class instances for training from the set of proposals containing less than 30\% of an object ground-truth box. To avoid biasing our networks, we assign uniformly distributed random poses to these, and assign depths as the centroid of points in the window. Over the course of training, proportion of ``none'' exemplars used was gradually reduced to help with pose and depth estimation performance for the other classes. Additionally, we experimented with training using a loss function specific to only one task (class, pose or depth) after training on the full combined task, but found no benefit to doing so - the specialized loss function (and gradients computed from it) did not allow the models to increase their performance in the selected task in a significant way.

Training times ranged from approximately 12 hours for the simpler models to up to 48 hours for the most complex model on a Titan X GPU. For the largest model we were constrained by memory (12Gb) to using a relatively small batch size - we would expect better performance with larger batches.

  
\subsection{Transfer to Real Data}
In order to improve performance on the NYUv2 dataset \cite{Silberman12}, we use transfer learning to adapt our synthetically trained networks to the new, more difficult, domain. We experimented with three strategies for adaptation: 1. Only allow the high-level layers in the network to adapt, keeping the two lowest-level layers fixed as a ``feature-extractor'' (we choose two layers based on \cite{Yosiniki14}), 2. Allow all levels of the network to adapt, and 3. Alternate training iterations between full-adaptation iterations and iterations on synthetic data, with the proportion of synthetic data being reduced over the course of training. To avoid over-fitting as well as to allow adaptation of the none-class, we use bounding box proposals for training (in addition to the ground truth boxes).


\section{Experimental Evaluation}
In this section we evaluate the performance of our trained models on each sub-task independently. We show results for both our synthetic test set as well as on the NYUv2 dataset of Silberman \etal \cite{Silberman12}. When possible, we compare to the state of the art, and show how our method both outperforms and is subject to fewer constraints - primarily because we train directly on cluttered noisy data. We also evaluate our different network architectures on our own synthetic test set. Finally, we present qualitative results which show the ability of our method to provide semantic understanding and pose for full scenes. 

\subsection{Architectures}
\begin{figure*}[t]
  \centering
  \includegraphics[width=1\textwidth]{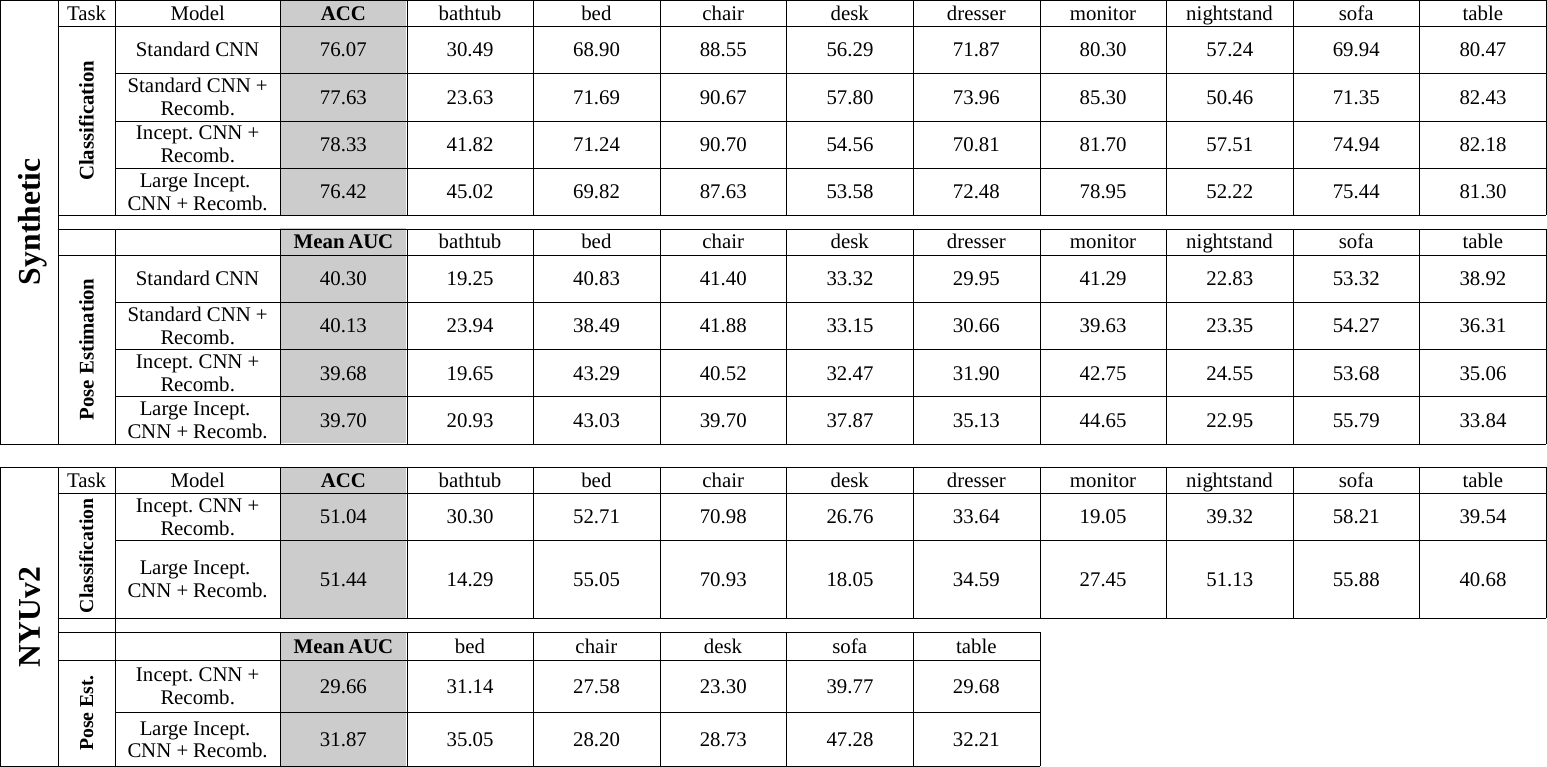}
  \caption{Performance of the four CNN architectures on synthetic (top) and NYUv2 (bottom) datasets for classification and pose estimation. Per class classification results show the F1-score, per class pose estimation results show the normalized AUC measure. See Secs. \ref{seg:eval_class} and \ref{seg:eval_pose} for details. All numbers are in percent.}
  \label{fig:class_pose_table}
\end{figure*}

We first compare results from our four different network architectures of increasing complexity. In Fig.\ref{fig:class_pose_table} we give per-category and averaged results for all four models on all three sub-tasks. As can be seen, the difference between the models is not very substantial - leading us to believe that we were actually limited by the size of our training set. This seems to be confirmed by the fact that the larger two networks began over-fitting towards the end of their training runs.

\subsection{Classification}
\label{seg:eval_class}
To evaluate classification performance, we classify the ground truth bounding boxes from our synthetic test set and the NYUv2 dataset. Fig.~\ref{fig:class_pose_table} gives per class accuracy on both datasets. As an overall measurement for the classification we use accuracy (ACC) being the fraction of correctly classified samples across all classes. Per class performance is measured using the F1-score as the harmonic mean of recall and precision.
Unfortunately, while we would like to compare to other works, each recent work has reported classification accuracy slightly differently. Lin \etal \cite{Lin13} merge similar classes (\eg table and desk), and only report an overall number. Couprie \etal \cite{Couprie13} only report pixel-wise accuracy, which we do not compute, as we do not need such a fine-grained segmentation. Gupta \etal \cite{Gupta15} do not evaluate their classification independently and instead give detector average precision (AP).

\subsection{Pose}
\label{seg:eval_pose}
To measure absolute pose estimation performance, we evaluate the estimated pose per class against ground truth poses. As all objects are located on the floor plane (or in the case of monitors, a horizontal supporting surface), we need only estimate a rotation around the floor normal. For our synthetic set we compare against the ground truth poses used to render the data, while for the NYUv2 dataset we use the annotations of Guo and Hoeiem \cite{Guo13}. We only include results for the 5 of our trained classes for which Guo and Hoeiem provided annotation (bed, chair, desk, sofa, and table). For both synthetic and real datasets, we use the ground truth boxes as our input to isolate pose estimation performance.

\begin{figure*}[t]
  \centering
  \includegraphics[width=1.0\textwidth]{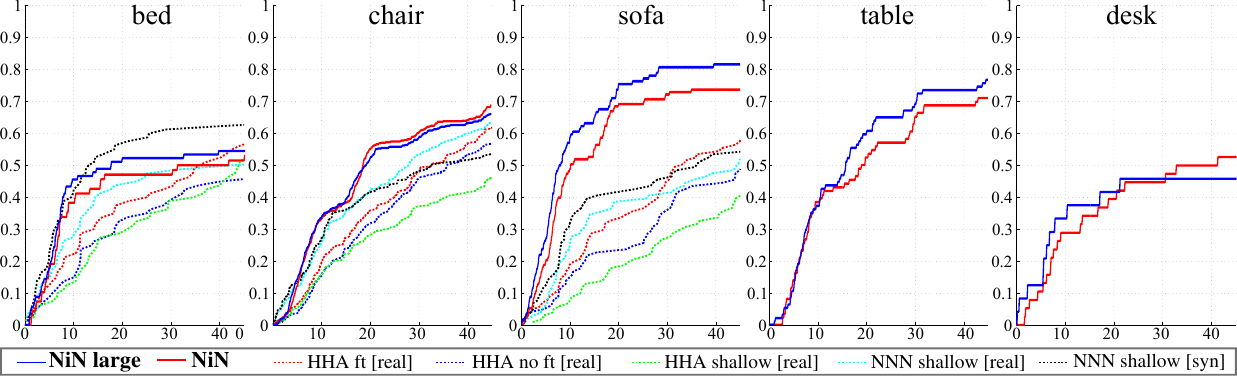}
  \caption{Pose estimation performance five classes in the NYUv2 \cite{Silberman12} test set. We plot accuracy versus allowed angular error $\delta_\theta$. Our methods (NiN and NiN large - solid lines) outperform the state of the art results of Gupta \etal \cite{Gupta15}.}
  \label{fig:pose_angular_error}
\end{figure*}

To compute a real valued pose and depth we extract the value of the maximum bin and its two neighbors, and compute a weighted sum using the bin centers, \ie
\begin{equation}
 \theta = {\sum_{i=-1}^{1} \theta_{hist}(\kappa+i)*\theta_{\kappa+i}  \over \sum_{i=-1}^{1} \theta_{hist}(\kappa+i) }: \kappa = \argmax_k {\theta_{hist}(k)},
\end{equation}
where $\theta_{\kappa}$ is the angle at the center of bin $\kappa$, and $\theta_{hist}(\kappa)$ is the soft-maxed value of bin $\kappa$. We only consider the local distribution around the max bin so that our estimates are not corrupted by multi-peaked histogram distributions (which occur due to rotational symmetries). We should also note that 90 degree rotational symmetries are an unavoidable source of error for some of our classes, especially tables and night stands. To evaluate pose error, we adopt the measure of \cite{Gupta15}, which plots the accuracy vs increasing allowed angular error $\delta_\theta$. To retrieve a scalar performance measure for the pose estimation we use a normalized Area-Under-Curve (AUC) for threshold values up to 15 degrees. For overall performance we average the values weighted by number of instances per class. As seen in Fig.~\ref{fig:pose_angular_error}, we strongly outperform the state of the art \cite{Gupta15} in two classes, with slightly poorer performance in the other.

\subsection{Qualitative Results}
\begin{figure*}[t]
  \centering
  \includegraphics[width=1.0\textwidth]{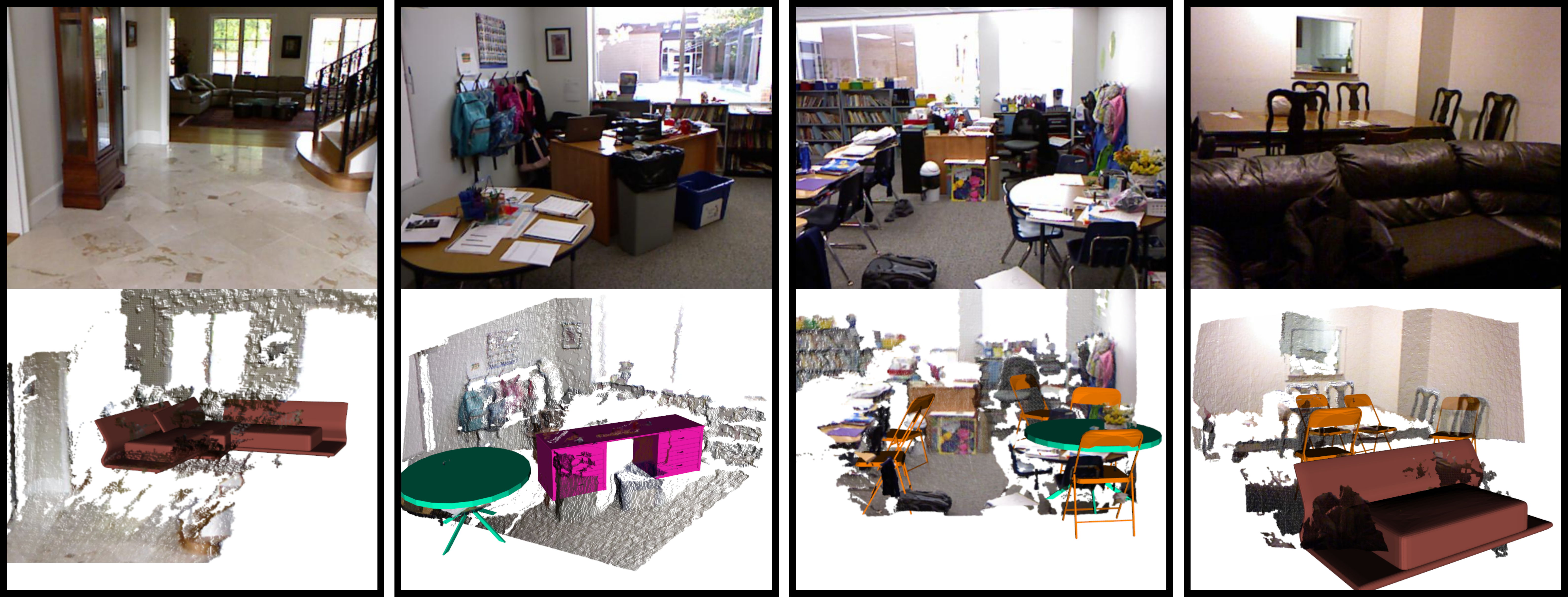}
  \caption{Qualitative pose and classification results on the NYUv2 dataset. Within each pair: Top: Original Scene; Bottom: Point-cloud with generic pose-aligned models inserted. Note that classification and pose-estimation are model independent. For visualization we used a random model per class.}
  \label{fig:qualitative_nyu}
\end{figure*}

To combine our classifier results, we first use non-maximum suppression (NMS) to disentangle and remove multiple detections with an allowed overlap of 20\%. Then we combine all bounding box activations using a per-pixel max-pooling scheme. Figure~\ref{fig:qualitative_nyu} shows an example of the semantic heatmaps generated this way, which give a rough class-wise labeling of the scene. Additionally, we show placed generic 3D models for each detected object to show results of pose estimation.

\section{Conclusions}
We have presented a method for generating realistic synthetic RGB-D scenes for training vision algorithms to segment, classify, and estimate pose and position of common furniture classes. We then showed that these scenes can be used to train deep CNNs to recognize and estimate pose for objects of the classes trained on, even if the object models tested on were not part of the training set; that is, the networks can be used to solve class-based pose estimation, rather than specific model-based pose as has been the prevailing standard. 

Furthermore, we have also demonstrated with several experiments that networks trained on synthetic RGB-D scenes can be adapted easily to work on the most challenging real data available, even if the amount of annotated real data available is relatively small. Moreover, we have accomplished all three tasks within a single network, allowing understanding of full scenes in a matter of seconds on a modern GPU. Furthermore, with our pipeline the amount of training data available is practically limitless, as we generate the next batch while the current scenes are trained on - the only limitation is the number and types of models. Future work will expand the pose-aligned classes to include the full ModelNet40 dataset and should add further cues to generate even more realistic procedural scenes. We expect this to allow future researchers to extend even further the complexity and performance of machine learning techniques on RGB-D data.

{\small
\bibliographystyle{ieee}
\bibliography{iccv2015}
}

\end{document}